\documentclass[a4paper]{article}
\usepackage{iwslt18,amssymb,amsmath,epsfig}
\usepackage{times}
\usepackage{latexsym}
\usepackage{amsmath}
\usepackage{url}
\usepackage{mathtools}
\usepackage{CJKutf8}
\usepackage{multirow}
\usepackage{footnote}
\usepackage{footmisc}

\DeclarePairedDelimiter{\floor}{\lfloor}{\rfloor}
\def\reg{{\rm\ooalign{\hfil
     \raise.07ex\hbox{\scriptsize R}\hfil\crcr\mathhexbox20D}}}

\renewcommand{\thefootnote}{\alph{footnote}}

\newcommand{\astfootnote}[1]{
\let\oldthefootnote=\thefootnote
\setcounter{footnote}{0}
\renewcommand{\thefootnote}{\fnsymbol{footnote}}
\footnote{#1}
\let\thefootnote=\oldthefootnote
}

\title{Word-based Domain Adaptation for Neural Machine Translation}

\makeatletter
\def\name#1{\gdef\@name{#1\\}}
\makeatother
\name{{\em Shen Yan}, {\em Leonard Dahlmann}, {\em Pavel Petrushkov}, {\em Sanjika Hewavitharana}, {\em Shahram Khadivi}}
\address{
eBay Inc.  \\
{\small \tt \{shenyan,fdahlmann,ppetrushkov,shewavitharana,skhadivi\}@ebay.com
}}
\begin{document}
\maketitle
\begin{abstract}
In this paper, we empirically investigate applying word-level weights to adapt neural machine translation to e-commerce domains, where small e-commerce datasets and large out-of-domain datasets are available. In order to mine in-domain like words in the out-of-domain datasets, we compute word weights by using a domain-specific and a non-domain-specific language model followed by smoothing and binary quantization. The baseline model is trained on mixed in-domain and out-of-domain datasets. Experimental results on En $\rightarrow$ Zh e-commerce domain translation show that compared to continuing training without word weights, it improves MT quality by up to $2.11\%$ BLEU absolute and $1.59\%$ TER. We have also trained models using fine-tuning on the in-domain data. Pre-training a model with word weights improves fine-tuning up to $1.24\%$ BLEU absolute and $1.64\%$ TER, respectively.
\end{abstract}

\section{Introduction} \label{sec_1}
Domain adaptation (DA) techniques in machine translation (MT) have been widely studied. For statistical machine translation (SMT), several DA methods have been proposed to overcome the lack of domain-specific data. For example, self-training \cite{Ueffing2005,Schwenk2008} uses a MT system trained on general corpus to translate in-domain monolingual data as additional training sentences. Topic-based DA \cite{Tam2007, Hewavitharana2013} employs topic-based translation models to adapt for different scenarios. Data selection approaches \cite{Moore2010IntelligentSO, Axelrod2011DomainAV, Duh2013AdaptationDS, Vogel2015UsingJM} first score the out-of-domain data using language model trained on both domain-specific and non-domain-specific monolingual corpora, then rank and select the out-of-domain data that are similar to in-domain data. Instance weighting methods \cite{Matsoukas2009DiscriminativeCW, Foster2010DiscriminativeIW} score each sentence/domain using statistical rules, then train the MT models by giving sentence/domain-level scores.   

Neural machine translation (NMT) has become state-of-the-art in recent years \cite{Sutskever:2014:SSL:2969033.2969173, DBLP:journals/corr/BahdanauCB14, Luong2015StanfordNM, Wu2016, Vaswani2017AttentionIA}. There are several research works on NMT domain adaptation. For example, back-translation methods \cite{Sennrich2016ImprovingNM} use a NMT model trained on the reverse direction to translate domain-specific monolingual data as additional training sentences. Fast DA approaches \cite{Luong2015StanfordNM, Freitag2016FastDA} train a base model using mixed in-domain and out-of-domain datasets, then fine-tuning on in-domain datasets. Mixed fine-tuning \cite{Chu2017AnEC} combines fine-tuning and multi-domain NMT.  Similar to instance weighting in SMT, sentence/domain weighting methods \cite{Chen2017, Wang2017} can also be used for NMT domain adaptation based on different objectives. DA with meta information \cite{Khadivi2017NeuralAS} is proposed to train topic-aware models using domain-specific tags for the decoder. Chunk weighting method \cite{Petrushkov2018LearningFC} describes a way of selecting and integrating positive partial feedback from model-generated sentences into NMT training.       

In this paper, we propose word-level weighting for NMT domain adaptation. We compute the word weights in out-of-domain datasets based on the logarithm difference of probability according to a domain-specific language model and non-domain-specific language model followed by smoothing and binary quantization. This gives the in-domain words in out-of-domain sentences higher weights and biases the NMT model to generate more in-domain-like words. Thus, the work presented in this paper can be viewed as a generalization of instance weighting. To remove noise in the word weights, we study the effectiveness of using smoothing methods. Specifically, a weighted moving average filter is proposed to apply smoothing to the computed word scores with its nearby words.

Experiments on En $\rightarrow$ Zh e-commerce domain translations tasks show that: 1) Domain adapted model with smoothed word weights gains significant improvement over non-smoothed weights; 2) Continuing training the model with computed word weights improves translation results significantly compared to continuing training without word weights; and 3) Compared to directly fine-tuning on in-domain datasets, fine-tuning after pre-training with word weights results in translation performance improvement on the in-domain e-commerce test set.            

The rest of the paper is structured as follows. The approach and model we use is described in Section 2, where we
first recap the NMT objective and then present the details of the proposed word-level weighting approach. Experimental results and discussions are presented in Section 3 and Section 4, followed by conclusions and outlook in Section 5.

\section{Approach} \label{sec_2}
We present word weighting objective on NMT before discussing how to generate the weights. 
\subsection{Objective}
In this work we use attention-based neural machine translation model \cite{Sutskever:2014:SSL:2969033.2969173,DBLP:journals/corr/BahdanauCB14, Wu2016} for experiments. Given a parallel bilingual dataset $\textit{D}$, the NMT model is trained to maximize the conditional likelihood \textit{L} of a target sequence \emph{$y_1^T$ : $y_1$ , $\ldots$ , $y_T$} given a source sequence \emph{$x_1^N$ : $x_1$ , $\ldots$ , $x_N$}:

\begin{equation} \label{eq_1}
\textit{L} = \sum_{(x_1^N, y_1^T) \in \textit{D}}\sum_{t=1}^T \log{p(y_t|y_1^{t-1}, x_1^N)} 
\end{equation}

Training objective (\ref{eq_1}) can be simply modified to word-level loss \textit{$L_w$} with word weights $w_t$:

\begin{equation} \label{eq_2}
\textit{$L_w$} = \sum_{(x_1^N, y_1^T, w_1^T) \in \textit{D}}\sum_{t=1}^T w_t\log{p(y_t|y_1^{t-1}, x_1^N)} 
\end{equation}

The word weights $w_t$ for a target sequence $y_1^T$ can be 0 or 1. We set $w_t$ = 1 for all in-domain sentences. For out-of-domain sentences, $w_t$ = 1 means the word in the out-of-domain sentence is related to in-domain datasets (selected), $w_t$ = 0 means it is not. 

Our training objective (\ref{eq_2}) can be seen as a generalization of the original training objective (\ref{eq_1}) and instance weighting methods \cite{Chen2017, Wang2017}. The original loss (\ref{eq_1}) sets $w_t$ = 1 for every word in all sentences. The instance-level loss can be expressed as giving a target sentence, $w_t = w \ \forall t$, where w is the weight for the sentence or the domain. Our training objective is similar to \cite{Petrushkov2018LearningFC}, however, instead of generating chunk-based user feedback for model predictions, we compute the word weights using language models trained on real target data.    

\subsection{Approaches to the objective} \label{approach}
To compute discriminative word weights, we first follow the data selection methods in SMT \cite{Moore2010IntelligentSO}. To state this formally, let $I$ be the domain-specific corpus, $O$ be the non-domain-specific corpus, and $y_t$ be the word in out-of-domain sentences at target position $t$. We denote by $P_I(y_t | y_{t-n}^{t-1})$ the per-word probability conditioned on previous $n-1$ words, according to a language model trained on $I$. Similarly, we denote by $P_O(y_t| y_{t-n}^{t-1})$ the per-word probability conditioned on previous $n-1$ words according to a language model trained on $O$. We can estimate $P_I(y_t | y_{t-n}^{t-1})$ and $P_O(y_t| y_{t-n}^{t-1})$ by training language models on $I$ and $O$, separately. Therefore, the word scores $s_t$ can be computed in the log domain: 

\begin{equation} \label{eq_4}
\textit{$s_t$} =  \log{P_I(y_t | y_{t-n}^{t-1})} -  \log{P_O(y_t | y_{t-n}^{t-1})} 
\end{equation}

Since the value of $s_t$ is strongly correlated with the neighborhood words, it is worth investigating smoothing of the word scores before binary thresholding to remove the noise. Hence, a weighted moving average kernel:

\begin{equation} \label{eq_5}
\textit{$\hat{s_t}$} = \sum_{k=\floor{\frac{-L}{2}}}^{\floor{\frac{L}{2}}}c_{k}s_{t+k} 
\end{equation}

is then applied to smooth word score $s_t$ at each target position $t$. Here $L$ is the kernel size and $c_k$ are values of the kernel for $k \in [\frac{-L}{2}, \frac{L}{2}]$.  
In our experiments, we heuristically set the values of the kernel based on mean average with $c_k$ = $c$ = $\frac{1}{L}$ or gaussian distribution with $c_k$ = $\frac{1}{\sqrt{2\pi}\sigma}e^{\frac{-k^2}{2{\sigma}^2}}$, where we set $\sigma$ to be the global variance of the word scores.  

The special case of sentence-level weights can be expressed as $\hat{s_t} = \hat{s} \ \forall t$, where $\hat{s}$ is the averaged smoothed word scores for the target sentence $y_1^T$. In this case, the training objective (\ref{eq_2}) becomes equivalent to sentence weighting method from \cite{Wang2017} with appropriately modified scoring function.

After smoothing the word scores, we finally binarize the smoothed word scores based on a threshold $T$:

\begin{equation}
    w_t= 
\begin{cases}
    1,& \text{if } \hat{s_t} \geq T\\
    0,              & \text{otherwise}
\end{cases} 
\end{equation}

In our experiments we set the threshold $T=0.5$ and only keep the words above the threshold. This means we select a word if $w_t=1$ and do not select it if $w_t=0$. Considering word weights $w_t$ are gathered in a binary form during continuing training, the selected words would be good candidates that we want to extract from out-of-domain corpus $O$. In fact, word weights $w_t$ are precomputed offline and used during the training. It can be set to any real value, depending on the way of thresholding.

\subsection{Chunk-based weighting} \label{LCW}
\begin{figure*}[hpt]
\centering
\includegraphics[width=1.0\textwidth]{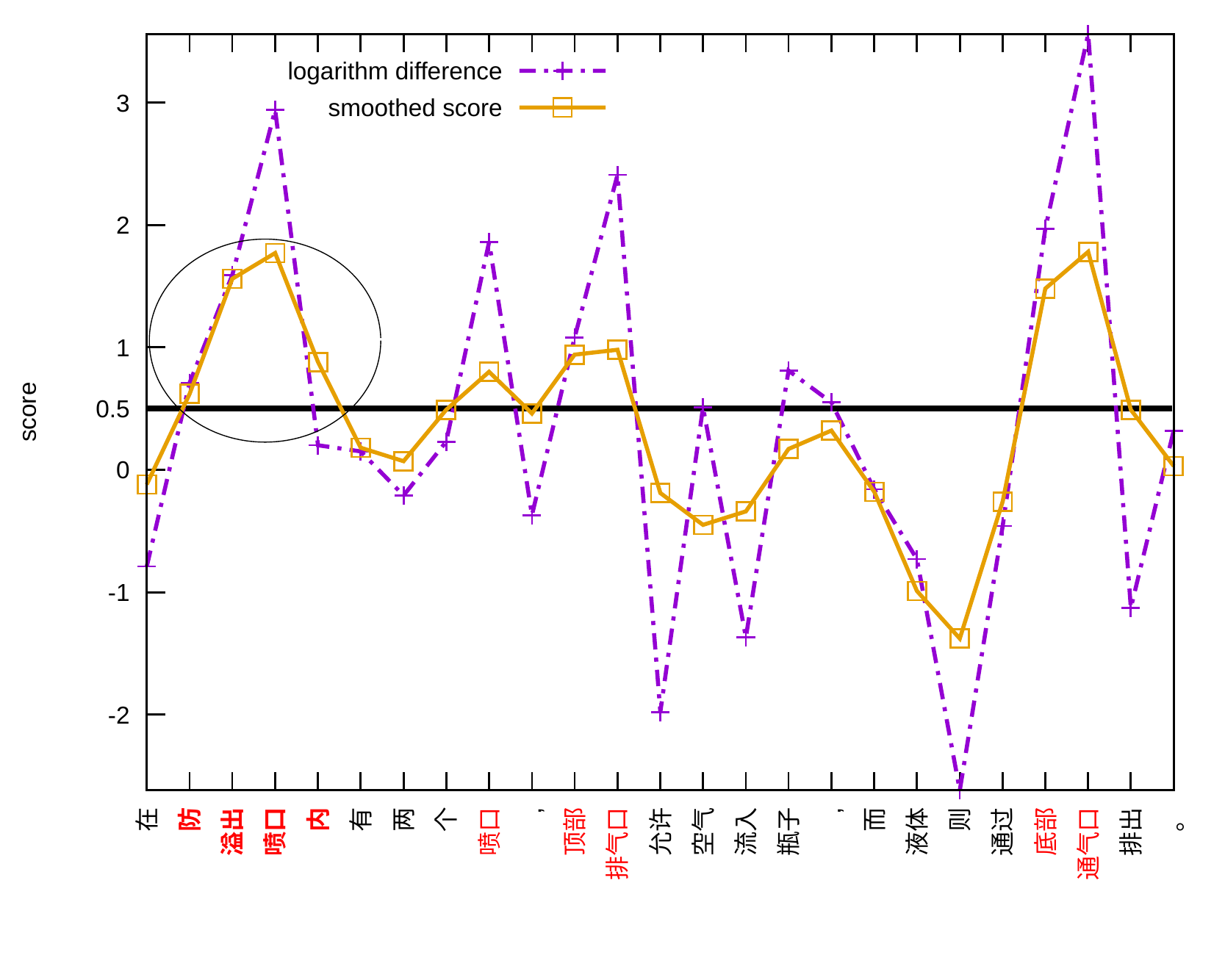}
\caption{Our approach for word-based domain adaptation, with the target sentence on the bottom (the source sentence is "Inside the non-spills spout there are two vents, where the top vent allows air to flow into the bottle while liquid smoothly pours out through the bottom vent."). Above are displayed the word scores and the smoothed word scores. The black bold line indicates the threshold and the circle indicates which chunk is selected by LCW. After smoothing, the isolated random words are removed  and the red words on the bottom are selected. The red bold words on the bottom are preserved after LCW.}
\label{fig_1}
\end{figure*}

Considering that the selected words in a target sentence might still be noisy and we select single random words, we alternatively experimented with selecting only the part (chunk) in the target sentence that has the longest consecutive weights (LCW) with $w_t=1$. For each target sentence, we pick only one chunk and set all other weights to zero. See Figure \ref{fig_1} for an example. Then, because the surrounding context is also selected, the chunk is less likely to be noise. If there are multiple such chunks with the same length in the sentence, we simply randomly sample one of them. We found that the chunk-based approach in practice performs slightly better than word-level weighting.

\section{Experiments} \label{sec_3}
In this section, we conduct a series of experiments to study how well NMT performs when word-level weights are given for out-of-domain training data. We also study the effectiveness of the smoothing methods.    

\subsection{Datasets and data processing} \label{data processing}
We report the results on our in-house English-to-Chinese e-commerce item descriptions dataset. Item descriptions are provided by private sellers and like any user-generated content, may contain
ungrammatical sentences, spelling errors, and other type of noise. We first segmented the Chinese sentences with Stanford Chinese word segmentation tool \cite{Chang2008OptimizingCW} and tokenized English sentences with the scripts provided in Moses \cite{Koehn2007MosesOS}. On both languages, we use subword units based on byte-pair encoding (BPE) \cite{Sennrich2016bpe} with 42,000 subword symbols learned separately for each language. For En-Zh we have $0.53$M in-domain e-commerce sentence pairs and $5.15$M sampled out-of-domain sentence pairs (UN, subtitles, TAUS data collections, etc.) that have significant n-gram overlap with the item description data. We validate our models on an in-house development set consisting of $3173$ item descriptions, and evaluate on an in-house test set of $739$ item descriptions using case-insensitive character-level BLEU \cite{Papineni2002BleuAM} and TER \cite{Snover2006ASO} with in-house tools. For development and test sets, a single reference translation is used. Statistics of the data sets are reported in Table \ref{table_1}.

\begin{table*}[t]
\centering
\begin{tabular}{|c|c|c|c|}
\hline
\multicolumn{2}{|c|}{Data set}                     & \multicolumn{2}{c|}{e-commerce + out-of-domain} \\ \hline
\multicolumn{2}{|c|}{Language}                     & English          & Chinese           \\ \hline
\multirow{3}{*}{Training} & Sentences              & \multicolumn{2}{|c|}{5,689,989} \\ \cline{2-4} 
                          & Running words          & 97,266,344          & 96,480,106          \\ \cline{2-4} 
                          & BPE vocabulary        &  33,484           & 45,867           \\ \cline{1-4} 
\multirow{2}{*}{Dev}      & Sentences              & \multicolumn{2}{|c|}{3173 (item descriptions)}             \\ \cline{2-4} 
                          & Running words          & 51,130            & 48,900           \\ \hline 
\multirow{2}{*}{Test}     & Sentences              & \multicolumn{2}{|c|}{739 (item descriptions)}            \\ \cline{2-4} 
                          & Running words          & 19,034            & 18,262 \\ 
\hline
\end{tabular}
\caption{Corpus statistics for the e-commerce English$\rightarrow$Chinese MT tasks.}
\label{table_1}
\end{table*}

To compute our word weights we train a domain-specific 4-gram language model and a non-domain specific 4-gram language model using KenLM \cite{Heafield2011KenLMFA}. For the domain-specific language model, we collected domain-specific monolingual data from an e-commerce website, resulting in the number of $15$M sentences. For the non-domain-specific language model, we use sampled LDC Chinese Gigaword (LDC Catalog No.: LDC2003T09) with $36$M sentences. It should be noted that we train our language models on the word-level. In order to score a BPE-level corpus with such a language model, we score its words and copy this score for each of the subword units. After the word scores are computed, we then smooth them via a guassian distributed kernel with window size $L=5$. We choose window size $L=5$ considering that the language model is trained based on sequences of four words. We observed similar results with different window sizes, which is discussed in Section \ref{sec_4}. Finally, we binarize the smoothed word scores into binary word weights by setting the threshold $T=0.5$. The computed word weights are applied to the target side of out-of-domain sentences during the phase of continuing training. In order to get better translation results, we first trained the baseline model with mixed in-domain and out-of-domain data according to training objective \ref{eq_1}, where no weights are used. We start our experiments by continuing training from this baseline model.

We implemented our NMT model using Tensorflow \cite{Abadi2016TensorFlowAS} library. The encoder is a bidirectional LSTM with size of 512 and the decoder is a LSTM with 2 layers of same size. All the weight parameters are initialized uniformly in $[-0.1, 0.1]$. We set dropout on RNN inputs with dropping probability $0.2$. We train the networks with batch size $120$ using SGD with initial learning rate $1.0$ and gradually decaying to $0.1$ after the initial $2$ epochs.  

\subsection{Results}

\begin{table}[t]
\begin{center}
\begin{tabular}{|l|l|l|}  \hline
Corpus & Sent. count & Token count \\ \hline
ood. sentences & 5,153,191 & 93,427,867 \\ \hline
+sent. weights & 2,633,109 & 26,275,096 \\ \hline
+word weights & 3,873,264 & 36,617,395  \\ \hline
+chunk weights & 3,873,264 & 25,813,480 \\ \hline
\end{tabular}
\end{center}
\caption{\label{table_2} Out-of-domain training corpus statistics. \emph{ood. sentences} indicates the number of sentences/tokens in the out-of-domain corpus. \emph{+sent. weights} indicates the number of selected out-of-domain sentences where the weights of the sentences are equal to $1$. \emph{+word weights} and \emph{+chunk weights} indicate the statistics of selected out-of-domain sentences/tokens  after applying word weights generation and LCW methods as described in Section \ref{approach} and \ref{LCW}.} 
\end{table}

Statistics of the out-of-domain sentences/tokens selection after applying different types of weights are summarized in Table \ref{table_2}. Before the selection, the number of out-of-domain sentences is $5.15$M and the number of tokens is $93.4$M. When sentence-level weights are used, the sentences with $w_t=0$ are ignored, resulting in the number of remaining sentences/tokens around $2.63$M and $26.3$M, respectively. When word-level weights are used, there are $1,279,927$ sentences where all word weights in the sentences are equal to zero. After removing these sentences, around $3.87$M sentences are preserved and the number of selected tokens with word weights $w_t$=$1$ is around $36.6$M. Given computed word weights, we alternatively choose only the chunk with the longest consecutive weights (LCW) where $w_t=1$, resulting in chunk-level weights with the selected number of tokens further reduced to $25.8$M.         

We train a baseline NMT model on mixed in-domain and out-of-domain data with objective defined as Eq. \ref{eq_1} for $6$ epochs. The data is mixed completely (mixed $0.53$M in-domain e-commerce and $5.15$M sampled out-of-domain sentence pairs) while training the baseline model. The baseline model initialized by a mix of in-domain/out-of-domain data can be regarded as a kind of "warm start". We have also tried training a baseline with out-of-domain data only and observed slightly worse result after fine-tuning on in-domain data (0.5 BLEU). Hence, we use the baseline model trained on a mix of in-domain/out-of-domain data in the following experiments. Given the baseline model, we then directly fine-tune on in-domain data for another $10$ epochs or first continue training on the mixed data with sentence/chunk/word weights for $3$ epochs and then fine-tune on in-domain data for $10$ epochs. The model is saved after each epoch. We take the model which gives the best result on our development set for evaluation. Note that we always set word weights $w_t=1$ for our in-domain dataset.

\begin{table*}
\centering
\begin{tabular}{|l|c|c|c|}
\hline
\multicolumn{2}{|c|}{}   & \multicolumn{2}{c|}{Item descriptions}  \\
\hline
No. & System description & BLEU [\%] & TER [\%]  \\
\hline
1 & Baseline & 24.37 & 61.66 \\
\hline
2 & 1 + continue training without word weights    & 24.31 & 61.69 \\
3 & 1 + continue training with sentence weights & 25.79 & 60.82 \\
4 & 1 + continue training with word weights & 26.14 & 60.34 \\
5 & 1 + continue training with chunk weights & 26.42 & 60.10 \\
\hline
6 & 1 + fine-tuning on in-domain  & 26.06 & 59.93  \\
7 & 5 + fine-tuning on in-domain & 27.30 & 58.29 \\
\hline
\end{tabular}
\caption{\label{table_3} E-commerce English $\rightarrow$ Chinese BLEU results on test set. \emph{Baseline} is trained on mixed in-domain and out-of-domain data.  \emph{No. 2} is continuing training from baseline with objective defined as Eq. \ref{eq_1}. \emph{No. 3} is continuing training from baseline with sentence-level weights and \emph{No. 4} is with word weights, as defined in Section \ref{approach}. \emph{No. 5} refers to assigning $w_t$ using LCW method described in Section \ref{LCW}. \emph{No. 6} is equivalent to directly fine-tuning on in-domain datasets starting from the baseline model and \emph{No. 7} is equivalent to fine-tuning on in-domain datasets after \emph{No. 5} is finished.}
\end{table*}

In Table \ref{table_3}, we show the effect of different types of weights on translation performance. First, the baseline trained on mixed in-domain and out-of-domain datasets gives $24.37\%$ BLEU and $61.66\%$ TER, respectively. Directly fine-tuning on in-domain dataset already improves the model due to the bias of the model towards in-domain data. 

Continuing training on mixed datasets with previous objective defined in Eq. \ref{eq_1} shows insignificant changes in terms of BLEU and TER. However, introducing sentence-level weights improves the model from $24.37\%$ to $25.79\%$ BLEU and $61.66\%$ to $60.82\%$ TER, respectively. Compared to continuing training without weights, sentence-level weights are generated as described in Section \ref{approach}, where $w_t \ \forall t$ are set to the same sentence weight $w \in \{0, 1\}$ . We set the threshold equal to $0.5$ and keep the sentences with weights above the threshold. The result from sentence-level feedback suggests that mining good out-of-domain sentences which are similar to in-domain datasets and dissimilar to out-of-domain datasets benefits model translation towards in-domain-like sentences even without fine-tuning on in-domain datasets.

The use of word-level weights improves the baseline model even better, from $24.37\%$ to $26.14\%$ BLEU and $61.66\%$ to $60.34\%$ TER, respectively. In this approach, the number of selected tokens is drastically reduced to $36.6$M from $93.4$M tokens, nearly $61\%$ drop in number of tokens with improved translation performance.  Word-level weights also outperform sentence-level weights by $0.35\%$ in BLEU score and $0.48\%$ in TER. It can be explained by the fact that each word in the sentences are given its own similarity to the in-domain datasets. Considering sentence-level weights set all words in a sentence with the same weight, even though part of the words in the sentences might not be related to the in-domain corpus, word-level weights are more accurate and effective.   

Finally, chunk-level weights are generated from our word-level weights based on LCW. Here we aim to train the domain-adapted model from more consecutive segments rather than single selected words. On top of word-level weights, it improves by another $0.28\%$ BLEU absolute and $0.24\%$ TER, respectively. Out-of-domain sentences can be split into chunks which can be related to the in-domain and can be translated independently in terms of the context. The selection of the consecutive chunk with in-domain-like context can positively affect the training towards domain-adapted model. By focusing on in-domain related and out-of-domain unrelated part, word/chunk-level weights can effectively reduce the unnecessary noise in the out-of-domain training data. Compared to continuing training without word weights, we are able to further reduce the corpus by $72\%$ tokens ($25.8$M vs. $93.4$M selected tokens), resulting in an improvement of $2.11\%$ BLEU absolute and $1.59\%$ TER, respectively. It should also be noted that with similar number of tokens ($25.8$M  vs. $26.3$M), chunk-level weights outperforms sentence-level weights by $0.63\%$ BLEU absolute and $0.72\%$ TER.

Next, we further fine-tune the model with chunk-level weights and obtain further improvements of $0.88\%$ BLEU absolute and $1.81\%$ TER.  Compared to directly fine-tuning on the baseline, continuing training the model with chunk-level weights and then fine-tuning improves translation results from $26.06\%$ to $27.30\%$ BLEU and $59.93\%$ to $58.29\%$ TER, respectively.

\begin{table}
\begin{center}
\begin{tabular}{|l|c|c|} \hline
 System & BLEU [\%] & TER [\%]  \\ \hline 
 Baseline  & 24.37 & 61.66 \\ \hline
 +w.w. without smooth. & 21.38 & 66.25 \\ \hline
 +w.w. (mean smooth.)   & 25.99 & 60.70 \\ \hline
 +w.w. (gauss. smooth.)  & 26.14 & 60.34 \\ \hline
\end{tabular}
\end{center}
\caption{\label{table_4} Study on the effect of different smoothing methods for word weights generation. \emph{Baseline} is the same as before. \emph{w.w. without smoothing} means the word weights (w.w.) are computed without smoothing in the log domain. \emph{w.w. (mean smooth.)} indicates smoothing the word scores via using a mean average filter before thresholding and \emph{w.w. (gauss. smooth.)} indicates using a normal distributed filter before thresholding. The approaches regarding different smoothing methods are described in Section \ref{approach}.}
\end{table}

Results from the study on the effect of using different smoothing methods are shown in Table \ref{table_4}. The word weights generated without using smoothing methods, where $\hat{s_t} = s_t$, lead to poor translation quality of $21.38\%$ from $24.37\%$ BLEU and $66.25\%$ from $61.66\%$ TER, respectively. We need to smooth the word scores before thresholding because the values of $\log{P_I(y_t | y_{t-n}^{t-1})} -  \log{P_O(y_t | y_{t-n}^{t-1})}$ are noisy. If there are selected isolated words like '$,$' which have higher scores than the surrounding text, it may cause rare vocabulary problem after training.

The results from word weights computed from mean averaged filter and normal distributed filter are relatively close, $25.99\%$ vs. $26.14\%$ BLEU and $60.70\%$ vs. $60.34\%$ TER, respectively. These results are obtained via a filter with window size $L=5$. In practice, we also tried setting window size $L=3$ and $L=7$, but didn't observe different results. We found that the surrounding word scores have to be considered for smoothing in order to make the word weights $w_t$ less noisy as well as more precisely representing the similarity to the in-domain/out-of-domain. 

Additionally, we also experimented with randomly selecting words in the out-of-domain sentences with binary mask. However, we observed a drop in the translation accuracy.

\subsection{Examples}
In Table \ref{table_example}, we show an example for which the system trained with word weights produces a better translation. The English sentence is "non-spill spout with patented valve".
The word "spout" is rare in our data, appearing in the out-of-domain training sentences only once. The Chinese side of this training example can be seen in Figure \ref{fig_1} together with the weights assigned to the individual words by our method. When smoothing is applied, isolated Chinese words such as \begin{CJK*}{UTF8}{gbsn} "空气" \end{CJK*} ("air") are removed. With the longest consecutive words (LCW) method, the only remaining chunk is \begin{CJK*}{UTF8}{gbsn} "防/溢出/喷口/内" \end{CJK*} ("inside the non-spills spout"), which is related to our in-domain data. The system with word weights is then trained only on this chunk on the target side, while the baseline model is trained on the entire sentence and generates inappropriate translations.


\section{Discussions} \label{sec_4}
The domain adaptation techniques (sentence-level/chunk-level/word-level) introduced in this paper are all derived from word weights generation. They aim to select out-of-domain sentences/chunks/words which are more related to in-domain corpus and unrelated to out-of-domain corpus. The word weights are computed prior to system tuning via the logarithm difference of LM probability scoring and are then used for tuning the sequence-to-sequence model.  By measuring domain similarity with external criteria such as LM, this kind of out-of-domain data selection is able to highlight the in-domain-related and out-of-domain-unrelated parts and leads to less variation and errors in our e-commerce domain adaptation. In addition, the selected out-of-domain segments have to be smoothed in order to reduce noise.

\begin{table*}[hpt]
\begin{center}
\begin{tabular}{|l|l|} \hline
 System & Translation  \\ \hline 
 Baseline & \begin{CJK*}{UTF8}{gbsn} 带 专 利 阀 的 防 溢 出 溅 漏 \end{CJK*} \\ \hline
 + word weights & \begin{CJK*}{UTF8}{gbsn} 带 专 利 阀 的 防 溢 出 喷 口 \end{CJK*} \\ \hline
 Reference & \begin{CJK*}{UTF8}{gbsn} 带 有 专 利 阀 门 的 防 溢 口 \end{CJK*} \\ \hline
\end{tabular}
\end{center}
\caption{\label{table_example} Translations of "non-spill spout with patented valve" produced by the baseline NMT system and the system trained with word weights. The last row shows the reference translation.}
\end{table*}

\section{Conclusions} \label{sec_5}
In this work, we generate word-level weights by calculating the logarithm difference of the probability of two external language models for domain adaptation. This approach better selects the out-of-domain segments related to e-commerce domain, and requires fewer tokens for training. We experimented with continuing training models with sentence/chunk/word weights and show that they all give translation improvement in terms of BLEU and TER compared to continuing training without word weights. Experiments on our in-house English-Chinese datasets also show that continuing training with word weights then fine-tuning improves results over directly fine-tuning on baseline model.

In future, with the computed word weights as the initial parameters, we want to devise strategies to make online domain adaptation possible by dynamically updating word weights during training, which could in turn lead the in-domain data translation to better match its references.

\bibliographystyle{IEEEtran}
\bibliography{iwslt2018}
\end{document}